\documentclass[sigplan,accepted]{acmart}

\AtBeginDocument{%
  
}

\usepackage[utf8]{inputenc} 
\usepackage[T1]{fontenc}    
\usepackage{hyperref}       
\usepackage{amsmath}
\usepackage{cleveref}
\AtBeginEnvironment{appendices}{\crefalias{section}{appendix}}
\AtBeginEnvironment{appendices}{\crefalias{subsection}{appendix}}
\usepackage{url}            
\usepackage{booktabs}       
\usepackage{amsfonts}       
\usepackage{nicefrac}       
\usepackage{microtype}      
\usepackage{xcolor}         
\usepackage{wrapfig}

\usepackage{booktabs}
\usepackage{subcaption}
\usepackage{graphicx, wrapfig}
\usepackage{caption}
\usepackage{tablefootnote}
\usepackage{xspace}
\usepackage[title,titletoc]{appendix}
\usepackage{multirow}
\usepackage{adjustbox}
\usepackage{csquotes}
\usepackage{balance}
\usepackage{subcaption}
\usepackage[flushleft]{threeparttable}

\usepackage{tikz}
\def\checkmark{\tikz\fill[scale=0.4](0,.35) -- (.25,0) -- (1,.7) -- (.25,.15) -- cycle;}
\newcommand{\DP}[0]{($\epsilon$~,~$\delta$)~-~DP }

\newcommand{\papername}[0]{FLEdge\xspace}

\newcommand{\revision}[1]{{#1}}

\setcopyright{acmlicensed}
\acmYear{2024}
\copyrightyear{2024}
\acmConference[MIDDLEWARE '24]{25th International Middleware Conference}{December 2--6, 2024}{Hong Kong, Hong Kong}
\acmBooktitle{25th International Middleware Conference (MIDDLEWARE '24), December 2--6, 2024, Hong Kong, Hong Kong}
\acmDOI{10.1145/3652892.3700751}
\acmISBN{979-8-4007-0623-3/24/12}

\begin{document}

\title{\papername: Benchmarking Federated Learning Applications in Edge Computing Systems}

\author{Herbert Woisetschl\"ager}
\email{h.woisetschlaeger@tum.de}
\orcid{0000-0001-9729-2895}
\affiliation{%
  \institution{Technical University of Munich}
  \country{Germany}
}

\author{Alexander Erben}
\email{alex.erben@tum.de}
\affiliation{%
  \institution{Technical University of Munich}
  \country{Germany}
}

\author{Ruben Mayer}
\email{ruben.mayer@uni-bayreuth.de}
\affiliation{%
  \institution{University of Bayreuth}
  \country{Germany}
}

\author{Shiqiang Wang}
\email{wangshiq@us.ibm.com}
\affiliation{%
  \institution{IBM Research}
  \country{United States}
}

\author{Hans-Arno Jacobsen}
\email{jacobsen@eecg.toronto.edu}
\affiliation{%
  \institution{University of Toronto}
  \country{Canada}
}

\renewcommand{\shortauthors}{Woisetschl\"ager et al.}

\begin{abstract}
    \revision{
    Federated Learning (FL) has become a viable technique for realizing privacy-enhancing distributed deep learning on the network edge. 
    Heterogeneous hardware, unreliable client devices, and energy constraints often characterize edge computing systems.
    In this paper, we propose FLEdge, which complements existing FL benchmarks by enabling a systematic evaluation of client capabilities. 
    We focus on computational and communication bottlenecks, client behavior, and data security implications.
    Our experiments with models varying from 14K to 80M trainable parameters are carried out on dedicated hardware with emulated network characteristics and client behavior.
    We find that state-of-the-art embedded hardware has significant memory bottlenecks, leading to $4\times$ longer processing times than on modern data center GPUs.
    } 

\end{abstract}



\maketitle

    \section{Introduction}
    \label{sec:intro}
    Federated Learning (FL) has become an established middleware abstraction to facilitate distributed and privacy-preserving deep learning~(DL)~\cite{caldas1, He2020, wang2023flint}. 
With increasing access to data by eliminating the need to transfer data to a central location, FL reduces network communication and fosters privacy. 
Privacy in FL applications is typically realized by introducing differential privacy \cite{mcmahan2018dp, andrew2019}.
There is ample research on benchmarking FL workloads, which focuses on different fields of application, data heterogeneity~\cite{caldas1, He2020}, FL aggregation strategies~\cite{He2020, Lai2021}, DL model personalization~\cite{chen2022}, and federated hyperparameter optimization~\cite{Wang2022}. 
However, these benchmarks typically target server-grade hardware and are simulation-based \cite{caldas1, He2020}. As such, an often neglected factor is the type of system FL workloads are deployed to: edge computing systems.

Edge computing is a technology for managing a fleet of devices distributed across geographies~\cite{Varghese2016}, serving latency-critical or data-sensitive tasks~\cite{He2017, Ren2019}. Systems at the edge are characterized by diverse hardware and low client reliability~\cite{Cooke2020}, varying network quality~\cite{Li2018, Goudarzi2022}, and energy constraints~\cite{Jiang2020}. Additionally, FL workloads are prone to a high degree of data heterogeneity~\cite{Mitra2021, Ranzato2021, Fang2022}. This creates a set of challenges when placing FL workloads at the edge since servers and clients are strongly intertwined. While clients facilitate the training on their local data, servers use FL strategies (e.g., FedAvg) to aggregate the trained client models into a global model that generalizes over all clients. 

As clients in edge computing systems are often embedded devices, their computational capabilities and points of most energy-efficient operations vary. Additionally, devices can be more than 5 years old in current systems that require high functional safety~\cite{dell_edge, siemens_edge}. Older devices usually do not have DL accelerating hardware, while more recently introduced platforms are powerful system-on-a-chip (SoC) devices carrying a GPU~\cite{orin2022, roadmap_jetson_2020}, leading to a major gap in computational capabilities. In practice, this requires careful deployment planning for FL workloads, especially when large language models (LLMs) are involved~\cite{Desislavov2023}.

\revision{
FLEdge extends existing FL benchmarking works by introducing systematic studies on client behavior, communication and energy efficiency, and hardware diversity. 
As these dimensions are critical to FL system design, we help researchers and practitioners develop new FL efficiency methods and drive practical adoption of FL, especially in edge computing systems.

}

\textbf{Client behavior}. State-of-the-art benchmarks for FL applications provide a variety of DL models, datasets, and FL strategies \cite{Xie2022_FederatedScope, He2020, caldas1}. Yet, no work systematically studies the effects of client dropouts on the overall training performance in real systems. Client dropouts will likely happen in edge computing systems where clients and network connectivity can be unstable~\cite{Varghese2022}, which makes it an integral part to analyze, especially with regard to the interference effects with differential privacy (DP) that depend on the number of clients actually submitting model updates during a training round.

\textbf{Communication efficiency}. Equally important to the client behavior are network conditions, as it is an integral component for an efficient edge deployment. However, many contributions discuss communication efficiency in FL systems \cite{mcmahan2017, konecny2017, Sattler2020, Wu2022} without providing metrics that help estimate the viability of a workload to be federated into an edge computing system with wireless communication, like 4G \cite{3gpp_4g_lte}.

\textbf{Energy efficiency}. 
Existing benchmark works predominantly focus on computational speed~\cite{Xie2022_FederatedScope, Lai2021}.
However, two key design variables in edge computing systems are computational and energy efficiency, which both contribute to the overall energy efficiency of an FL system.
Energy is a scarce resource in edge computing systems and ultimately determines how fast we can process a workload depending on client hardware.

\textbf{Hardware diversity}. While simulated embedded devices can be used to evaluate the scalability of FL algorithms in depth, they neglect the underlying processor architecture and memory bandwidth. This results in significant performance differences when deploying on different types of hardware and limits the applicability to edge computing systems~\cite{beutel2020, He2020, Varghese2016}.

\revision{
    Overall, FLEdge aims to improve the understanding of FL application deployment in edge computing systems and specifically answers the following research question: 
}
 
\begin{displayquote}
    \textit{How well do state-of-the-art FL workloads respond to deployment in edge computing systems?}
\end{displayquote}
\noindent Our contributions are as follows:
\begingroup
    \renewcommand\labelenumi{(\theenumi)}
    \begin{enumerate}
        \item \textbf{We introduce \papername{} -- a hardware-centric benchmarking suite for FL workloads in system-hetero-geneous environments}. FLEdge extends the widely used FL framework Flower~\cite{beutel2020} with a module to control client behavior, an adaptive user-level differential privacy adapter that accounts for client dropouts, a network emulator, and extensive monitoring capabilities to evaluate the computational efficiency of embedded devices. Our code base is publicly available.\footnote{\revision{\url{https://github.com/laminair/FLEdge}}.}

        \item \textbf{\papername provides a holistic evaluation pipeline for FL workloads w.r.t. client behavior and communication efficiency}. The client behavior module allows flexible modeling of client reliability based on environmental conditions and the intended deployment target. This enables practitioners to quickly evaluate the robustness of their FL workloads. We also add a freely controllable network adapter that allows for the emulation of realistic connectivity and the exploration of the viability of deploying FL workloads in edge computing systems.

        \item \textbf{We conduct extensive experiments on computational capabilities and energy efficiency of widely used embedded devices when running FL workloads}. 
        Our experimental results show that the state-of-the-art embedded AI accelerator is challenged with backpropagation, which calls for alternative solutions that better use the hardware architecture. 
        Our experiments on client behavior in conjunction with differentially private FL workloads show a high sensitivity of model quality w.r.t. client reliability.
      
    \end{enumerate}
\endgroup

Our work is structured as follows. In \Cref{sec:background}, we outline the requirements analysis for our benchmark.
In \Cref{sec:methodology}, we introduce our methodology. In \Cref{sec:bm-design}, we outline our experimental design, including datasets, DL models, FL strategies, and practical considerations. \Cref{sec:evaluation} discusses experimental results of our benchmark. \Cref{sec:related-work} contains related work. \Cref{sec:lessons-learned} discusses lessons learned and in \Cref{sec:discussion-conclusion}, we conclude our work.

    \section{Requirement Analysis}
    \label{sec:background}
    As edge computing applications are gaining popularity, especially in realistic environments without perfect control over system parameters, challenges in designing efficient systems arise.
On a client, we care about energy efficiency and training reliability, as well as privacy, such that the data is processed in a timely and secure manner. 
At the same time, there are often (mobile) clients with heterogeneous computing resources involved, which may negatively affect the training speed. 
That said, we need a hardware-centric benchmarking framework to assess the feasibility of FL applications on the network edge.
Furthermore, as FL aims to be a communication-efficient ML paradigm, cost and efficiency are key decision variables. 
The objective of our work is to evaluate the end-to-end FL pipeline, investigating system components that have come short in existing research:

\textbf{Client behavior}. Generally, client reliability is key for a distributed application. While for most edge computing applications, tasks run individually on clients, e.g., in the context of CV applications with MobileNet models \cite{howard2017}, for FL, there are significant dependencies between clients and the server.
FL typically involves an iterative training procedure where a set of $N$ clients trains a shared model over multiple rounds $t \in T$. In each round, a subset $M \in N$ clients is selected for training. Training an FL model involves aggregating multiple client updates into a single model: $w^{t+1} = \frac{1}{|M|}\sum_{m \in M} w_m^t$. For this, we use FL strategies like FedAvg \cite{mcmahan2017}. In edge computing, clients are unreliable compared to data center services and may show a high failure rate. Thus, $M$ can vary within a training round, and consequently, fewer client updates will be aggregated. As such, it is a requirement to test the robustness of state-of-the-art FL strategies with varying realistic client behavior typically found in edge computing systems.

\textbf{Privacy}. Another key concern for FL systems is data privacy, as clients are unwilling to share their data. FL already increases the level of data privacy to a certain extent~\cite{mcmahan2017} but does not provide a \emph{guarantee} for privacy to clients. Formal methods like $(\epsilon, \delta)$-~Differential Privacy (DP) can be applied to obtain such a guarantee.
DP can be applied on sample- or user-level, where the latter is of special interest for FL \cite{mcmahan2018dp, andrew2019}. As clients in FL systems only share their model updates, it is sufficient to apply DP to model updates and achieve the same guarantees as with sample-level DP \cite{wei2019}. The objective of user-level DP is to deny membership inference or gradient inversion attacks. This only works if the noise level is calculated appropriately based on the number of clients that actually submit a model update in an FL training round. As such, bringing together robustness and data privacy requires user-level DP to adapt to suddenly failing clients is a requirement.

\textbf{Communication efficiency}. Along with varying client behavior and privacy levels, communication is an integral part of ensuring a high service quality of FL workloads in edge computing systems. In many use cases, tasks are running independently on clients \cite{Varghese2022}. However, in FL systems, there is a strong coupling between clients and the server as the global model state is maintained on the server. Therefore, we are interested in the processing latency and every aspect of network communication for FL on the edge. Typically, there are three scenarios for deploying edge networks. 
The first and most reliable is a wired connection, often found in factories \cite{Capra2019}. 
The second is a high-bandwidth wireless connection, such as 4G LTE \cite{3gpp_4g_lte}, and the third is a low-bandwidth and high latency connection, often found in remote areas~\cite{Varghese2022, Jararweh2016}. 
The anticipation of network connectivity is key as to whether it makes sense to federate the training process for a given DL model. Interestingly, existing works measure network conditions in different ways. Some focus on available bandwidth \cite{Xu2021, niknam2019}, and others focus on the amount of data transmitted \cite{pmlr-v162-wang22y, pmlr-v162-wang22o}. As such, there is a necessity for a comprehensive and unifying metric that quickly answers the question of whether it is viable to deploy an FL workload into an edge computing system with heterogeneous network conditions based on the amount of time we use on computation vs. communication. 

\begin{figure*}[!ht]
    \centering
    \raisebox{-\height}{\includegraphics[width=\textwidth]{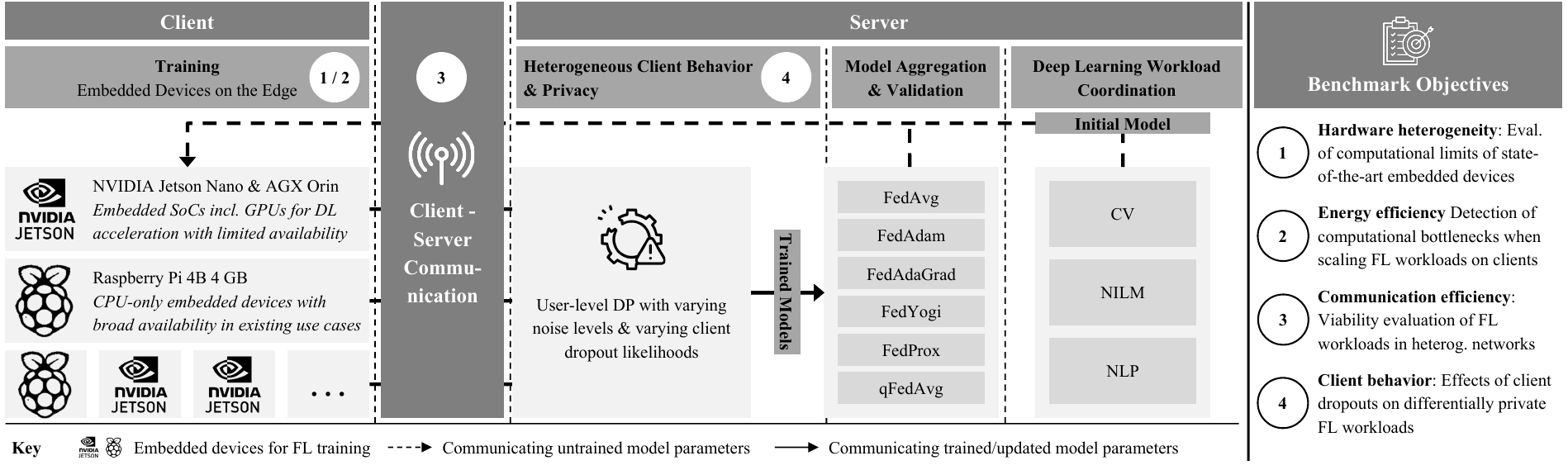}}
    \caption{Federated Learning protocol for one training round from a system perspective with 1 - 4 indicating the focus areas for our work and the benchmark subjects for FLEdge. Our system uses Flower as the underlying FL framework and implements each component in a modular and extensible manner.}
    \label{fig:system-design}
    \vspace{12pt}
\end{figure*}

\textbf{Communication cost}. 
FL applications in edge computing systems are often operated over wide-area networks \cite{Wang2019_survey}, involving a multitude of network hops.
While it is theoretically possible to measure and report the energy cost per hop, it is challenging to achieve in a real-world system since each networking component is often owned by a different entity, such as the client owner, the internet service provider, and the cloud operator.
Thus, we require a solution to reliably estimate the total communication cost of an FL workload.

\textbf{Energy efficiency}. To evaluate energy efficiency, we need a reliable method to evaluate the characteristics of existing devices for FL workloads. In edge deployments, clients have previously been employed as a data collection platform only and are now required to run computationally intensive FL workloads. This substantially increases energy consumption, which is often a challenge due to limited power availability \cite{Chen2019_energy}. To mitigate these effects, hardware acceleration on embedded devices has been introduced \cite{roadmap_jetson_2020}. 
For instance, GPUs generally provide a significant performance benefit over CPUs while being more energy efficient and cost-saving at the same time \cite{Steinkraus2005}. Therefore, it is a requirement to investigate the energy efficiency of different hardware w.r.t. their projected workloads. 
While DL workloads in the cloud are hard to measure for their energy efficiency due to virtualized hardware \cite{Fieni2021}, the benefit of embedded devices is their SoC design that allows for measurement of the real-time energy consumption of the entire device and all individual device components (e.g., CPU, GPU).

\textbf{Hardware diversity}. Edge computing is designed around the pattern of offloading computational tasks to edge devices, usually embedded devices in the proximity of a data source. This provides better control over the trade-off between computation and communication \cite{Mao2017}. 
Yet, a major challenge in these systems is hardware diversity, either due to long product life cycles in industrial systems or due to a lack of infrastructure control. 
For instance, widely used embedded devices for industrial edge require extensive reliability testing and therefore include hardware as old as five years~\cite{siemens_edge}. 
As such, it is necessary to benchmark not only the most recently released devices but also older generations still being used in a wide variety of systems~\cite{Wang2022_heterogen}.
Also, embedded devices are becoming increasingly capable as modern platforms, such as the NVIDIA Jetson AGX Orin, often entail integrated GPUs. This makes benchmarking hardware more complex and is also likely to change the energy demand of devices for workloads. A hardware-centric benchmark must, therefore, account for a variable number of components on an SoC-based embedded device.

    \section{FLEdge: Benchmarking Framework}
    \label{sec:methodology}
    \textbf{Client behavior}. As robustness is important to all FL workloads, and edge computing systems suffer from heterogeneous client behavior, we configure our testbed to simulate realistic client failures. For FL workloads, it does not matter whether the failure origin is hardware or communication. We model the client dropout as an independent binomial distribution since we assume clients will reenter training for future rounds and clients do not interfere with one another,
\begin{equation}
    p^d_m = P(m) = \mathrm{Bin}(p)\mathrm{.}
\end{equation}

We vary the failure likelihood $p$ from 0\% to 50\% in our experiments, with 0\% being very reliable clients as they are often found in factories and 50\% representing unreliable clients, e.g., mobile clients with wireless connetion. For example, the FL strategy FedAvg is extended with~$p^d_m$ as follows, 
\begin{equation}
    w^{t+1} = \frac{1}{|M|}\sum_{m \in M}(w_m^t \cdot p^d_m)\mathrm{.}
\end{equation}

The behavioral pattern can be freely exchanged for other patterns that suit particular use cases.

\textbf{Communication efficiency}. To transfer the model updates from the clients to the server, we need to consider communication efficiency and find a method to express the viability of adding an embedded computing platform under given network conditions.  
 To evaluate communication efficiency, we use granularity \cite{Hwang2010-cb}, which is an established metric for evaluating the training efficiency in a distributed system by comparing the computation time against the communication time,
 \vspace{-12pt}
 \begin{equation}
     G = \frac{T_{\mathrm{computation}}}{T_{\mathrm{communication}}}\mathrm{.}
 \end{equation}
 
$G \gg 1$ is considered favorable for distributing workloads, while $G \simeq 1$ or $G < 1$ indicates little utility when distributing a workload to a given system or device.
Our benchmark allows the emulation of different realistic connectivity profiles per client to explore the effect of $G$.

\textbf{Communication cost}. 
It is important to consider the communication costs of scaling an FL system since we operate in a data-parallel regime. 
While theoretically, we can scale the number of clients such that we fully saturate the server's network bandwidth, the effects of a large number of clients are limited \cite{charles2021on}. 
However, to establish the scalability-cost trade-off, we employ the \emph{per-bit communication cost model} that allows us to assert a client with constant communication costs per model update transmission \cite{Jalali2014}.
The model allows us to calculate the energy we consume at every hop in a network, 
\vspace{8pt}
\begin{equation}
\label{eq:communication_cost}
    \begin{aligned}
        P_{\mathrm{t}} = E_t \cdot \mathcal{B} =~& (n_\mathrm{as} \cdot E_\mathrm{as} + n_\mathrm{lc} \cdot E_\mathrm{lc} + n_\mathrm{lb} \cdot E_\mathrm{lb} + E_\mathrm{bng} \\ &+ n_e \cdot E_e 
        + n_c \cdot E_c + n_d \cdot E_d) \cdot \mathcal{B}\mathrm{.}
    \end{aligned}
\end{equation}
\vspace{8pt}

$E_\mathrm{as}$, $E_\mathrm{lc}$, $E_\mathrm{lb}$, $E_\mathrm{bng}$, $E_e$, $E_c$, $E_d$ are the per-bit energy consumption of $n_\mathrm{as}$ edge ethernet switches, zero or one $n_\mathrm{lc}$ LTE client modem, zero or one $n_\mathrm{lb}$ LTE base station, the broadband network gateway (BNG), one or more edge routers $n_e$, one or more core routers $n_c$, and one or more data center Ethernet switches $n_d$, respectively. $\mathcal{B}$ denotes the size of a model update in bits.

\textbf{Computational efficiency}. Aside from communication in edge computing systems, energy is a vital component of system efficiency as it may be a scarce resource in remote areas. Therefore, our testbed contains real-time measurement capabilities and control mechanisms to limit the power draw of each embedded device. This enables us to explore the deployment characteristics of FL workloads in a wide variety of edge computing systems.
We measure the energy efficiency as the throughput in samples per second (denoted as $Q$) divided by the average power draw (denoted as $W$) over the experiment time: 
\begin{equation}
    \eta_e = \frac{\mathrm{Q}}{W}\mathrm{.}
\end{equation}

\textbf{Hardware diversity}. To develop a detailed understanding of where potential inefficiencies could come from, we employ a micro-benchmark to study the effect of hardware diversity on the training performance of FL workloads. It focuses on the timing of the DL step times, namely the batch loading, forward, loss calculation, backward, and optimizer steps. With this, we get a detailed understanding of computational inefficiencies on embedded devices and uncover differences to data center hardware that potentially become bottlenecks for state-of-the-art FL workloads. 
\revision{Additionally, we use the PyTorch profiler with Kineto support to investigate performance bottlenecks in our FL clients. This provides us with the runtime of individual kernels, highlighting potential bottlenecks.}

\begin{figure}
    \centering
    \begin{subfigure}[!ht]{0.11\textwidth}
        \raisebox{-\height}{\includegraphics[height=2.8cm]{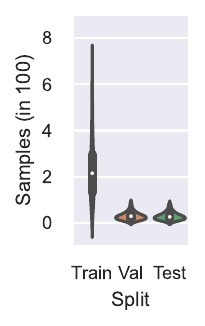}}
        \caption{BLOND}
        \label{fig:blond-dataset}
    \end{subfigure}
    \begin{subfigure}[!ht]{0.11\textwidth}
        \raisebox{-\height}{\includegraphics[height=2.8cm]{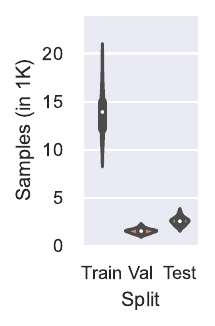}}
        \caption{FEMNIST}
        \label{fig:mnist-dataset}
    \end{subfigure}
    \begin{subfigure}[!ht]{0.11\textwidth}
        \raisebox{-\height}{\includegraphics[height=2.8cm]{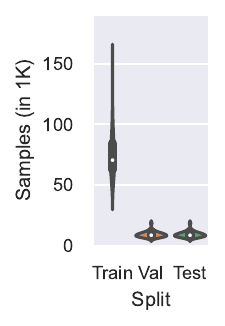}}
        \caption{Shakesp.}
        \label{fig:shake-dataset}
    \end{subfigure}
    \begin{subfigure}[!ht]{0.11\textwidth}
        \raisebox{-\height}{\includegraphics[height=2.8cm]{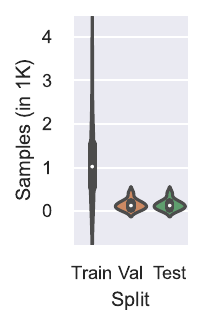}}
        \caption{Samsum}
        \label{fig:samsum-dataset}
    \end{subfigure}
    \caption{The non-IID subsets for our clients are sampled from a Dirichlet distribution ($\alpha= 1$). 
    }
    \label{fig:data-distributions}
\end{figure}

\subsection{Protocol}

\label{sec:fl-protocol}
Our setup follows the widely used client-server architecture for FL workloads \cite{caldas1, He2020, Xie2022_FederatedScope} and is depicted in \Cref{fig:system-design}. The four focus areas and their practical utility are discussed in the following.

\textbf{Training}. Training is facilitated entirely by the clients. They receive the hyperparameter configuration along with the model parameters from the server and train for exactly 1 local epoch before communicating the updates to the server. We do this for our client dropout and privacy experiments to isolate client failure effects on model quality and avoid any interference with data drift \cite{karimreddy2020scaffold, mohri2019-agnosticfl}.

\textbf{Communication}. We integrate a controllable network interface on each client as edge computing systems that involve embedded devices can be found in various environments (e.g., production lines or remote weather sensing stations). In addition to the steady 1 Gbit/s network link, which resembles industrial settings, our testbed allows us to emulate wireless communication (e.g., LTE, 3G) per client for remote setups.

\textbf{Privacy}. A core promise of FL is to preserve privacy. For user-level DP, the model aggregation method is extended by adding Gaussian noise to the model weights (cf. \Cref{eq:userlevel-dp}). With the additional noise $\xi \sim \mathcal{N}(0, I\sigma_{\Delta}^2)$, we introduce a natural trade-off between model accuracy and the level of privacy. The variance $\sigma_\Delta^2$ depends on the number of clients sampled in a training round and on their L2 update norm, as with more heterogeneous client updates, more noise has to be added to ensure \DP,
\begin{equation} 
    \label{eq:userlevel-dp}
    w^{t+1} = \frac{1}{|M|} \sum_{m \in M} ((w_m^t + z \cdot \xi) \cdot p^d_m)
\end{equation}

The strength of the privacy guarantee is measured by the privacy budget $\epsilon$ where lower values are better and provide higher levels of privacy \cite{Dwork2013}. $\epsilon$ depends on $\delta$, the likelihood of unintentional input data leakage. Usually, $\delta$ is set to a value equal to the inverse of the anticipated total number of samples in a dataset (e.g., for the BLOND dataset, it would amount to $\delta = \frac{1}{13,164}$, see \Cref{sec:datasets_models_strategies}).
As such, our system entails server-side user-level DP. In this way, we can adjust the DP noise levels on the server based on the number of updates received rather than having to request a partial re-calculation of DP model updates from each client.


\begin{table}[]
    \centering
    \caption{The datasets in our benchmark vary in modality, size, and data heterogeneity to resemble a large variety of real-world use cases.}
    \label{tab:pipelines}
    \resizebox{0.48\textwidth}{!}{
        \begin{tabular}{l|lrrlr}
\toprule
    Pipeline                & Dataset Name  & Total Samples   & Samples per Device  & Format    & Size \\ \midrule
    NILM                    & BLOND         & 13,164     & 234±132             & HDF5      & 5.46 GB \\
    CV                      & FEMNIST       & 814,255    & 13,958±2,106        & PNG       & 3.34 GB \\
    NLP                     & Shakespeare   & 4,226,054  & 75,131±18,281       & TXT       & 0.38 GB \\
    NLP                     & Samsum        & 11,780     & 491±723             & TXT       & 0.012 GB \\ \bottomrule \bottomrule
\end{tabular}
    }
\end{table}

\subsection{Practical Assumptions \& Configuration}
\label{subsec:prac-assumptions}

As we are interested in the performance of FL workloads, we configure our testbed to serve under realistic conditions for industrial edge computing systems. We describe the configuration along \Cref{fig:system-design}. 

\textbf{Training}. Clients are dedicated to the FL workload only and assign all available resources to the task with the respective optimal hyperparameter configuration as depicted in \Cref{tab:model-hparams}. 
We choose a client participation rate of $20\%$ for all experiments, but it can be freely configured for other scenarios.

\textbf{Communication}. The network is configured to either emulate a factory environment with a synchronous 1 Gbit/s link \cite{siemens_edge} or to run a client-specific 4G LTE wireless \cite{3gpp_4g_lte} network with higher latency and asynchronous 15 MBit/s upload and 40 MBit/s download bandwidths that is often found when using mobile or remotely embedded devices. The LTE bandwidths represent the global average connectivity \cite{trvisan2021_errant}. 
For the per-bit energy consumption model, we adopt the measurements and network topology from \citet{Jalali2014,Vishwanath2015} and consider two distinct scenarios for our communication cost analysis $P_t$: (I) a wired 1 Gbit/s interconnect as it can be found in, e.g., factories, and (II) a wireless setup where clients connect via an LTE gateway. For (I), we use a network topology with $n_\mathrm{as} = 2$, $n_\mathrm{lc} = 0$, $n_\mathrm{lb} = 0$, $n_\mathrm{bng} = 1$, $n_\mathrm{e} = 3$, $n_\mathrm{c} = 4$, $n_\mathrm{d} = 2$. For (II), we set $n_\mathrm{as} = 0$, $n_\mathrm{lc} = 1$, $n_\mathrm{lb} = 1$, $n_\mathrm{bng} = 1$, $n_\mathrm{e} = 4$, $n_\mathrm{c} = 4$, $n_\mathrm{d} = 2$. 

\textbf{Privacy}. The server is set up to handle both workloads that require very light privacy guarantees, i.e., for workloads that deal with non-sensitive data and for workloads ($\epsilon > 5$) that require very tight guarantees ($\epsilon < 1$) \cite{mcmahan2018dp, andrew2019}. 

\textbf{Client behavior}. The client dropout module is designed to account for a wide variety of use cases that create different wear and tear on embedded devices and reduce their reliability. Typically, in industrial environments, service quality is essential. Thus, the availability times often range well above 95\% of the operation time \cite{Ardagna2014}. With our study, we not only account for these services but also look into the effects of clients with very low reliability as they are found in loose collaborative learning tasks \cite{Hong2017}. To do so, we vary the client dropout likelihood from 0\% to 50\%. 

\textbf{Model aggregation \& validation}. The server runs state-of-the-art FL strategies and is configured according to related work. The detailed parameterization is available in \Cref{tab:fl-strategy-hparams}. 

\textbf{DL workload coordination}. The server is set up to coordinate state-of-the-art FL workloads that you would find in systems with both high and low reliability to provide a holistic picture of use cases.

\begin{table}[]
    \centering
    \caption{FL strategy hyperparameters. Key: $\eta$, $\eta_l$ = server-side learning rates, $\beta_1$, $\beta_2$ = Weight decay rates, $\tau$ = Server-side adaptivity level, $q$ = Fairness parameter for qFedAvg, $\mu$ = proximal parameter for FedProx.}
    \label{tab:fl-strategy-hparams}
    \resizebox{0.48\textwidth}{!}{
        \begin{tabular}{@{}ll|rrrrrrr|l@{}}
    \toprule
    FL Strategy              & Dataset     & $\eta$ & $\eta_l$ & $\beta_1$ & $\beta_2$ & $\tau$        & $q$        & $\mu$      & Further Details                       \\ \midrule
    FedAvg                   & All         & -      & -        & -        & -       & -           & -          & -          & FedAvg does not have                            \\ 
                             &             &        &          &          &         &             &            &            & any server hyperparameters                      \\ \midrule
    FedAdam                  & All         & $-1.5$ & $-1$     & $0.9$    & $0.99$  & $1e^{-2}$  & -          & -          &                                                 \\ \midrule
    FedAdaGrad               & All         & $0$    & $0$      & -        & -       & $1e^{-3}$  & -          & -          &                                                 \\ \midrule
    FedYogi                  & All         & $-1.5$ & $-1.5$   & $0.9$    & $0.99$  & $1e^{-5}$  & -          & -          &                                                 \\ \midrule
    \multirow{2}{*}{FedProx} & Shakespeare & -      & -        & -        & -       & -           & -          & $1e^{-2}$ &                                                 \\
                             & Others      & -      & -        & -        & -       & -           & -          & $1$        &                                                 \\ \midrule
    \multirow{2}{*}{qFedAvg} & Shakespeare & -      & -        & -        & -       & -           & $1e^{-2}$ & -          &                                                 \\
                             & Others      & -      & -        & -        & -       & -           & $1$        & -          &                                                 \\ \bottomrule \bottomrule
\end{tabular}
    }
\end{table}

    \section{Experimental Setup}
    \label{sec:bm-design}
    Our hardware-centric benchmark is deployed to a dedicated edge computing testbed. We use it to explore FL workloads in resource-constrained environments and explore the effects of power and network limitations on FL performance. For this benchmark, we deploy widely used state-of-the-art datasets from the NILM, CV, and NLP domains for FL applications, including an LLM workload.

\begin{table}
    \centering
    \caption{Our hyperparameters are chosen based on related work and hyperparameter sweeps for the best possible performance.
    }
    \label{tab:model-hparams}
    \resizebox{0.48\textwidth}{!}{
        \begin{tabular}{@{}ll|rrrrrr|r}
    \toprule
    Dataset                & Model      & Optim.    & LR    & W. Decay      & Dropout & Hidden Dim        & Params        & Minibatch         \\ \midrule
    \multirow{4}{*}{BLOND} & CNN        & SGD       & 0.055 & 0.0           &         &                   & 14,000        & 128               \\
                           & LSTM       & SGD       & 0.045 & 0.001         & 0       & 15                & 40,000        & 128               \\
                           & ResNet     & SGD       & 0.052 & 0.001         &         &                   & 100,000       & 128               \\
                           & DenseNet   & SGD       & 0.075 & 0.001         &         &                   & 252,000       & 128               \\ \midrule
    FEMNIST                & CNN        & Adam      & 0.001 & 0.0           &         &                   & 33,000        & 32                \\ \midrule
    Shakespeare            & LSTM       & SGD       & 0.8   & 0.0           & 0       & 256               & 819,000       & 32                \\ \midrule
    Samsum                 & FLAN-T5-Small & AdamW  & 0.0001& 0.0           & 0       & N/A               & 80,000,000    & 1 - 128           \\ \bottomrule \bottomrule
\end{tabular}

    }
\end{table}

\subsection{Testbed}
\label{sec:testbed}
We aim to explore FL applications in real systems by introducing two data center and three different embedded device types. 

\noindent\textbf{Data center GPU} (GPU). We use a GPU-accelerated data center node with 64 CPU cores, 256 GB of memory, and an NVIDIA A6000 (GPU) as our baseline device. The NVIDIA A6000 has a memory bandwidth of 768 GB/s. The VM has 3 TB NVMe storage. The VM is running Ubuntu 20.04 LTS with Python 3.9 and PyTorch 1.10. We use CUDA 11.6 and cuDNN 8.6.

\noindent\textbf{x86-CPU-base Clients} (VM). As a proxy for x86-based embedded devices without DL acceleration, we use virtualized systems with 4 CPU cores and 4 GB of memory (VM). The VM has a memory bandwidth of 25.6 GB/s and a network interconnect of 1 GBit/s. We use Ubuntu 20.04 LTS with Python 3.9 and PyTorch 1.10. We use the estimates from SelfWatts \cite{Fieni2021} as the power utilization profile for each VM. Fieni et al. calculate approx. 50 Watts for a 4 CPU core VM on Intel Xeon E5 processors.

\noindent\textbf{NVIDIA Jetson AGX Orin 64GB} (Orin). Our Orins are running Jetpack 5.1. This includes Ubuntu 20.04 LTS with Python 3.8, PyTorch 1.13, and CUDA 11.8. As the libraries are compiled specifically for the platform, we cannot adjust the stack to older PyTorch versions. The Orins have a memory bandwidth of 204 GB/s, a disk size of 64 GB, and a 10 Gbit/s network connection. They come with 2048 CUDA cores and 64 Tensor Cores. We measure power via the internal hardware-based monitoring functionality.

\noindent\textbf{NVIDIA Jetson Nano 2GB} (Nano). The devices run Ubuntu 18.04 LTS with Python 3.6 and PyTorch 1.10.\footnote{Jetson Nano with Maxwell architecture \cite{maxwell_aarch_2014} only supports CUDA 10.2 and PyTorch 1.10 with Python 3.6.} The Nano has a memory bandwidth of 25.6 GB/s. The latest supported CUDA version is 10.2 with cuDNN 7.2. The Nanos carry 128 CUDA cores but no Tensor cores. The maximum power draw of a Nano is 15 Watts.

\noindent\textbf{Raspberry Pi 4B 4GB} (RPi). They run with Ubuntu 20.04 LTS, Python 3.9, and PyTorch 1.10. There is no hardware acceleration available. The RPi has a memory bandwidth of 25.6 GB/s. The RPis have a class 10 32 GB SD card each and a 1GBit/s network interface. The RPis have a peak power draw of 10 Watts.

\vspace{-8pt}
\subsection{Software Stack}
\papername is implemented on top of widely used FL libraries. 
We use Flower \cite{beutel2020} to run the FL routine and PyTorch Lightning \begin{wraptable}[29]{r}[\dimexpr.75\width+.5\columnsep\relax]{10cm}
    \caption{Compounding effects of client dropouts and differential privacy for FedAvg across varying $z$ levels and client dropout rates.}
    \centering
    \label{tab:compound-effects}
    \resizebox{0.63\textwidth}{!}{
        \begin{tabular}{@{}cccccccccccc|cc@{}}
            \toprule
             & & \multicolumn{8}{c}{BLOND}& \multicolumn{2}{c}{FEMNIST} & \multicolumn{2}{c}{Shakespeare} \\ \midrule
             & \multicolumn{1}{c|}{} & \multicolumn{2}{c}{CNN} & \multicolumn{2}{c}{DenseNet} & \multicolumn{2}{c}{LSTM} & \multicolumn{2}{c|}{ResNet}& \multicolumn{2}{c|}{CNN} & \multicolumn{2}{c}{LSTM}\\
            $z$& \multicolumn{1}{c|}{$p$} & $\epsilon$& Acc.& $\epsilon$ & Acc.& $\epsilon$ & Acc.& $\epsilon$ & \multicolumn{1}{l|}{Acc.} & $\epsilon$& Acc.& $\epsilon$& Acc.\\ \midrule
            \multicolumn{2}{l|}{Loc. baseline} & N/A & 0.96& N/A& 0.89& N/A& 0.95& N/A& \multicolumn{1}{l|}{0.91} & N/A & 0.75& N/A & 0.59\\ \midrule
            \multirow{4}{*}{0} & \multicolumn{1}{c|}{0\%}& $\infty$ & 0.75& $\infty$& 0.74& $\infty$& 0.77& $\infty$& \multicolumn{1}{l|}{0.73} & $\infty$ & 0.70& $\infty$ & 0.53\\
             & \multicolumn{1}{c|}{10\%} & $\infty$ & 0.70& $\infty$& 0.73& $\infty$& 0.69& $\infty$& \multicolumn{1}{l|}{0.73} & $\infty$ & 0.69& $\infty$ & 0.52\\
             & \multicolumn{1}{c|}{20\%} & $\infty$ & 0.32& $\infty$& 0.69& $\infty$& 0.74& $\infty$& \multicolumn{1}{l|}{0.72} & $\infty$ & 0.68& $\infty$ & 0.51\\
             & \multicolumn{1}{c|}{50\%} & $\infty$ & 0.46& $\infty$& 0.67& $\infty$& 0.66& $\infty$& \multicolumn{1}{l|}{0.70} & $\infty$ & 0.66& $\infty$ & 0.49\\ \midrule
            \multirow{4}{*}{0.3} & \multicolumn{1}{c|}{0\%}& 8.0 & 0.73& 8.0& 0.70& 8.0& 0.74& 8.0& \multicolumn{1}{l|}{0.63} & 6.1 & 0.67& 6.6 & 0.53\\
             & \multicolumn{1}{c|}{10\%} & 8.1 & 0.73& 8.1& 0.70& 8.1& 0.70& 8.1& \multicolumn{1}{l|}{0.63} & 6.6 & 0.66& 6.6 & 0.52\\
             & \multicolumn{1}{c|}{20\%} & 8.1 & 0.40& 8.1& 0.69& 8.1& 0.70& 8.1& \multicolumn{1}{l|}{0.62} & 6.6 & 0.65& 6.6 & 0.49\\
             & \multicolumn{1}{c|}{50\%} & 8.1 & 0.44& 8.1& 0.67& 8.1& 0.64& 8.1& \multicolumn{1}{l|}{0.60} & 6.7 & 0.61& 6.6 & 0.44\\ \midrule
            \multirow{4}{*}{0.5} & \multicolumn{1}{c|}{0\%}& 2.3 & 0.54& 2.3& 0.64& 2.3& 0.74& 2.3& \multicolumn{1}{l|}{0.61} & 2.1 & 0.62& 2.0 & 0.52\\
             & \multicolumn{1}{c|}{10\%} & 2.4 & 0.54& 2.4& 0.63& 2.4& 0.74& 2.4& \multicolumn{1}{l|}{0.61} & 2.1 & 0.61& 2.0 & 0.49\\
             & \multicolumn{1}{c|}{20\%} & 2.4 & 0.38& 2.4& 0.60& 2.4& 0.72& 2.4& \multicolumn{1}{l|}{0.60} & 2.1 & 0.61& 2.0 & 0.45\\
             & \multicolumn{1}{c|}{50\%} & 2.4 & 0.36& 2.4& 0.57& 2.4& 0.70& 2.4& \multicolumn{1}{l|}{0.60} & 2.1 & 0.59& 2.0 & 0.44\\ \midrule
            \multirow{4}{*}{1} & \multicolumn{1}{c|}{0\%}& 0.4 & 0.51& 0.4& 0.56& 0.4& 0.64& 0.4& \multicolumn{1}{l|}{0.60} & 0.4 & 0.44& 0.4 & 0.52\\
             & \multicolumn{1}{c|}{10\%} & 0.4 & 0.51& 0.4& 0.56& 0.4& 0.61& 0.4& \multicolumn{1}{l|}{0.60} & 0.4 & 0.43& 0.4 & 0.44\\
             & \multicolumn{1}{c|}{20\%} & 0.4 & 0.36& 0.4& 0.56& 0.4& 0.60& 0.4& \multicolumn{1}{l|}{0.59} & 0.4 & 0.44& 0.4 & 0.44\\
             & \multicolumn{1}{c|}{50\%} & 0.4 & 0.34& 0.4& 0.54& 0.4& 0.57& 0.4& \multicolumn{1}{l|}{0.57} & 0.4 & 0.39& 0.5 & 0.39\\ \midrule
            \multirow{4}{*}{1.3} & \multicolumn{1}{c|}{0\%}& 0.2 & 0.48& 0.2& 0.56& 0.2& 0.69& 0.2& \multicolumn{1}{l|}{0.42} & 0.3 & 0.34& 0.3 & 0.51\\
             & \multicolumn{1}{c|}{10\%} & 0.2 & 0.45& 0.2& 0.56& 0.2& 0.64& 0.2& \multicolumn{1}{l|}{0.42} & 0.3 & 0.34& 0.3 & 0.44\\
             & \multicolumn{1}{c|}{20\%} & 0.2 & 0.38& 0.2& 0.56& 0.2& 0.63& 0.2& \multicolumn{1}{l|}{0.42} & 0.3 & 0.32& 0.3 & 0.44\\
             & \multicolumn{1}{c|}{50\%} & 0.2 & 0.34& 0.2& 0.51& 0.2& 0.61& 0.2& \multicolumn{1}{l|}{0.39} & 0.3 & 0.28& 0.3 & 0.39\\ \midrule
            \multirow{4}{*}{1.5} & \multicolumn{1}{c|}{0\%}& 0.2 & 0.34& 0.2& 0.30& 0.2& 0.37& 0.2& \multicolumn{1}{l|}{0.16} & 0.3 & 0.28& 0.3 & 0.48\\
             & \multicolumn{1}{c|}{10\%} & 0.2 & 0.34& 0.2& 0.30& 0.2& 0.37& 0.2& \multicolumn{1}{l|}{0.15} & 0.3 & 0.28& 0.3 & 0.44\\
             & \multicolumn{1}{c|}{20\%} & 0.2 & 0.37& 0.2& 0.29& 0.2& 0.35& 0.2& \multicolumn{1}{l|}{0.15} & 0.3 & 0.20& 0.3 & 0.43\\
             & \multicolumn{1}{c|}{50\%} & 0.2 & 0.19& 0.2& 0.28& 0.2& 0.31& 0.2& \multicolumn{1}{l|}{0.15} & 0.3 & 0.17& 0.3 & 0.36\\ \bottomrule
             \bottomrule
        \end{tabular}
    }
    \vfill
\end{wraptable} \noindent \cite{falcon2019pytorch} to allow for easy extensibility and seamless integration of new FL workloads. 

\subsection{FL Workloads}
\label{sec:datasets_models_strategies}

We use FL workloads that have been explored in previous benchmarking works.

\textbf{Datasets}. For NILM, we use the BLOND dataset \cite{Kriechbaumer2017}. It is an office environment appliance load monitoring dataset. It contains 13,164 samples with 12 appliance classes and captures electrical appliances in building-level office environments, such as laptops, monitors, and printers. We adopt Schwermer et al.'s \cite{Schwermer2022} approach to FL with BLOND. Each sample in the dataset consists of 25,600 power readings (current, voltage). 
Instead of generating the input features for the DL models in the data loader, we deviate from Schwermer et al. and create the Active Power, Apparent Power, Reactive Power, and MFCC (\texttt{n\_mfcc = 64}) input features offline \cite{barsim2014unsupervised}. This results in a reduced sample size of 68 x 1. The main reason for reducing the dataset size is to fit the dataset onto our RPi devices.

We employ the FEMNIST dataset \cite{caldas1} for CV. FEMNIST consists of 814,000 samples of hand-written digits and letters. The dataset contains 62 classes of handwritten digits (26 lower case letters, 26 upper case letters, and 10 digits). We randomly crop the grayscale samples to 28x28 and perform a random flip before training. Character recognition is frequently used by mobile clients to convert images to editable text documents.

We use the Shakespeare dataset \cite{He2020} for NLP. It consists of the complete works of William Shakespeare. It is preprocessed in the exact same way as introduced in the LEAF benchmark \cite{caldas1}. The dataset was preprocessed with a sliding window of 80 characters and a stride of 1 to prepare it for the next-character prediction task. The vocabulary was generated over the alphabet, including special characters, resulting in a total size of 80. Overall, the nature of the Shakespeare dataset resembles a task like a next-word prediction on smartphone keyboards.

We complement experiments in the NLP domain with the SAMSum dataset, which contains 16,000 pairs of chat-message-like dialogue and summary pairs that may be used for sequence-to-sequence modeling tasks \cite{gliwa-etal-2019-samsum}. With SAMSum, we introduce a realistic use case that can be used on mobile clients to provide a quick overview of their chat history, as it is frequently found in applications like Slack. 
Apart \newline\newline \vspace{28\baselineskip}\newline\noindent from splitting the dataset into 10 Dirichlet subsets ($\alpha = 1$), we do not apply additional preprocessing.
Researching FL applications on physical devices usually requires fitting the data distribution to the number of devices on our testbed.
We opt to sample client subsets to 45 based on a Dirichlet distribution ($\alpha = 1$) to fit the number of same-type clients in our testbed (Figure \ref{fig:data-distributions}).

\textbf{Models}. We train a total of 7 DL models. 
We use four different architectures to train on the BLOND datasets to showcase the sensitivity of different model sizes and architectures to real-world environmental conditions:  a CNN, an LSTM, a ResNet, and a DenseNet architecture \cite{Schwermer2022}.
To train on the FEMNIST dataset, we train a small and efficient CNN architecture, originally presented in the LEAF benchmark \cite{caldas1}. 
For the Shakespeare dataset, we use an LSTM model with 256 hidden dimensions and zero dropout that has been well explored to solve the next character prediction task \cite{He2020}.
Additionally, we use the SAMSum dataset to evaluate the hardware performance of the NVIDIA Jetson AGX Orin devices. To do so, we employ FLAN-T5-Small, an 80M parameter state-of-the-art transformer model \cite{chung2022flant5}. When fine-tuned for a specific downstream task, the FLAN-T5 model family has proven to deliver on-par performance with significantly larger foundation models such as LLama2 \cite{tinytitans_flant5}.
\begin{wraptable}[32]{r}[\dimexpr.75\width+.45\columnsep\relax]{11cm}
    \caption{Global model validation accuracy across FL strategies with varying client dropout rates after 100 FL training rounds with a client selection rate of 20\% per training round. \textbf{Bold} highlights the best-performing FL strategy.}
    \label{tab:dropout-analysis}
    \centering
    \scalebox{0.8}{
        \begin{tabular}{ll|rrrr|r|r}
            \toprule
            & \multicolumn{1}{c}{} & \multicolumn{4}{c}{BLOND}& \multicolumn{1}{c}{FEMNIST} & \multicolumn{1}{c}{Shakespeare} \\ \midrule
            $p$ & Strategy & \multicolumn{1}{c}{CNN} & \multicolumn{1}{c}{DenseNet} & \multicolumn{1}{c}{LSTM} & \multicolumn{1}{c|}{ResNet}& \multicolumn{1}{c|}{CNN} & \multicolumn{1}{c}{LSTM}\\
            \midrule
            \multicolumn{2}{l|}{\multirow[t]{6}{*}{\begin{tabular}{l}Loc. Baseline\\($p = 0\%$)\vspace{-10pt}\end{tabular}}} & 0.96±0.02 & 0.89±0.01 & 0.95±0.01 & 0.91±0.01 & 0.75±0.02 & 0.59±0.01 \\
            & & & & & & & \\
            \midrule
            \multirow[t]{6}{*}{0\%} & FedAdaGrad& 0.76±0.0& 0.17±0.07 & 0.71±0.03 & 0.70±0.0& 0.56±0.01 & 0.52±0.01 \\
            & FedAdam & 0.73±0.01 & 0.03±0.0& 0.64±0.05 & 0.64±0.02 & 0.38±0.05 & 0.5±0.03 \\
            & FedAvg& 0.75±0.02 & 0.74±0.0& 0.77±0.01 & 0.73±0.01 & \textbf{0.7±0.0}& \textbf{0.53±0.0} \\
            & FedProx & 0.75±0.01 & 0.74±0.0& 0.74±0.01 & 0.73±0.01 & \textbf{0.7±0.0}& 0.52±0.0 \\
            & FedYogi & \textbf{0.8±0.01} & \textbf{0.79±0.0} & \textbf{0.81±0.01}& \textbf{0.78±0.0} & 0.53±0.0& 0.25±0.0 \\
            & qFedAvg & 0.73±0.0& 0.7±0.0 & 0.79±0.0& 0.71±0.0& 0.03±0.0& 0.47±0.0 \\
            \midrule
            \multirow[t]{6}{*}{10\%}& FedAdaGrad& 0.76±0.03 & 0.03±0.0& 0.69±0.02 & 0.68±0.02 & 0.55±0.04 & 0.51±0.02 \\
            & FedAdam & 0.73±0.01 & 0.03±0.0& 0.54±0.1& 0.41±0.22 & 0.35±0.01 & 0.48±0.01 \\
            & FedAvg& 0.7±0.04& 0.73±0.01 & 0.69±0.04 & 0.72±0.0& \textbf{0.69±0.01}& \textbf{0.52±0.01} \\
            & FedProx & 0.74±0.05 & 0.73±0.01 & 0.68±0.03 & 0.73±0.01 & 0.68±0.01 & \textbf{0.52±0.01} \\
            & FedYogi & \textbf{0.76±0.02}& \textbf{0.78±0.01}& \textbf{0.8±0.0}& \textbf{0.76±0.03}& 0.53±0.01 & 0.23±0.01 \\
            & qFedAvg & 0.73±0.0& 0.69±0.0& 0.77±0.0& 0.71±0.0& 0.03±0.0& 0.47±0.0 \\
             \midrule
            \multirow[t]{6}{*}{20\%}& FedAdaGrad& 0.64±0.1& 0.04±0.0& 0.48±0.3& 0.69±0.03& 0.49±0.06 & 0.49±0.02 \\
            & FedAdam & 0.73±0.0& 0.03±0.0& 0.53±0.13 & 0.63±0.0& 0.34±0.05 & 0.35±0.06 \\
            & FedAvg& 0.32±0.27 & 0.69±0.05 & 0.74±0.03 & 0.72±0.01 & \textbf{0.68±0.02}& \textbf{0.51±0.02} \\
            & FedProx & 0.37±0.3& 0.69±0.05 & 0.68±0.04 & 0.72±0.01 & 0.67±0.01 & \textbf{0.51±0.02} \\
            & FedYogi & \textbf{0.76±0.02}& \textbf{0.76±0.03}& \textbf{0.76±0.0} & \textbf{0.75±0.03}& 0.47±0.08 & 0.21±0.03 \\
            & qFedAvg & 0.72±0.01 & 0.68±0.0& 0.75±0.0& 0.71±0.0& 0.03±0.0& 0.47±0.0 \\
             \midrule
            \multirow[t]{6}{*}{50\%}& FedAdaGrad& 0.42±0.24 & 0.03±0.0& 0.35±0.27 & 0.68±0.04 & 0.45±0.06 & 0.45±0.03 \\
            & FedAdam & 0.25±0.25 & 0.03±0.0& 0.18±0.13 & 0.62±0.01 & 0.2±0.09& 0.35±0.02 \\
            & FedAvg& 0.46±0.16 & 0.67±0.06 & 0.66±0.01 & 0.7±0.02& \textbf{0.66±0.03}& \textbf{0.49±0.03} \\
            & FedProx & 0.52±0.16 & 0.67±0.06 & 0.68±0.04 & 0.7±0.02& 0.64±0.02 & \textbf{0.49±0.03} \\
            & FedYogi & \textbf{0.72±0.01}& \textbf{0.71±0.09}& \textbf{0.74±0.04}& \textbf{0.74±0.04}& 0.43±0.12 & 0.2±0.04 \\
            & qFedAvg & 0.68±0.01 & 0.67±0.0& 0.72±0.01 & 0.71±0.0& 0.03±0.0& 0.47±0.0 \\
            \bottomrule
            \bottomrule
        \end{tabular}
    }
\end{wraptable}

 \vspace{-\baselineskip}\noindent
\noindent For the models used on BLOND, FEMNIST, and Shakespeare, we train one local epoch on clients and then send updates for aggregation to the server to eliminate potential risks of client drift~\cite{mcmahan2017}. 
The exact model configuration and hyperparameters are available from \Cref{tab:model-hparams}.

\textbf{FL strategies}. We explore all models in conjunction with six state-of-the-art FL strategies. We include FedAvg, the first communication efficient FL strategy, aggregating client updates over an unweighted average~\cite{mcmahan2017}. We also include adaptive strategies introduced by \citet{Reddi2020}, namely Fed-Adam, FedYogi, and Fed-AdaGrad. The adaptive strategies introduce a server-side learning rate to better account for data heterogeneity. We further adopt two strategies that aim to increase fairness and robustness in an FL system. FedProx~\cite{li_fedprox2018} introduces a method for weighting client updates in the aggregation process based on the amount of data a client has processed for an update. q-fair FedAvg (qFedAvg)~\cite{LiT2019} is a derivative of FedAvg introducing a factor $q$ that determines the level of fairness. Fairness in this context is defined by how well a model generalizes across clients. Higher generalizability is achieved by overweighting those clients that have the highest loss. This is done to gear a model stronger towards the high-loss clients and reduce the accuracy variance across clients. 
We fix the number of FL rounds to 10 for all FL strategies.
Further details on the datasets and DL models are available in \Cref*{tab:fl-strategy-hparams}.

\textbf{Network}. We modify our testbed communication to realistically reflect real-world scenarios with ERRANT~\cite{trvisan2021_errant}. We use a 1 GBit/s synchronous network link as well as the global average 4G LTE connection characteristics of 40 MBit/s download and 15 MBit/s upload~\cite{trvisan2021_errant}.


    \vspace{-8pt}
    \section{Results}
    \label{sec:evaluation}
    To show the practical utility of \papername, we run extensive experiments evaluating FL workloads on client behavior, differential privacy, energy efficiency, and hardware diversity.
\begin{table*}
    \caption{Computation and communication cost overview for our seven FL workloads across varying network conditions and 100 FL communication rounds (20\% selection rate, 9 training clients per round). Wireless communication drives communication costs by one order of magnitude.}
    \label{tab:granularity-network}
    \resizebox{\textwidth}{!}{
        \begin{tabular}{@{}lll|rr|rr|rr|rr|rr|lll|lll@{}}
        \toprule
        & && \multicolumn{4}{c|}{Communication}& \multicolumn{6}{c|}{Computation} & \multicolumn{6}{c}{$G$}\\ \midrule
        
        & && \multicolumn{2}{c|}{4G LTE}& \multicolumn{2}{c|}{1 GBit/s fiber} & \multicolumn{2}{c|}{RPi 4} & \multicolumn{2}{c|}{Nano} & \multicolumn{2}{c|}{Orin} & \multicolumn{3}{c|}{4G LTE}& \multicolumn{3}{c}{1 Gbit/s fiber} \\
        
        Dataset & Model & \multicolumn{1}{l|}{\begin{tabular}[l]{@{}l@{}}Communicable\\ parameters\end{tabular}} & \multicolumn{1}{c}{\begin{tabular}[c]{@{}c@{}}Time\end{tabular}} & \multicolumn{1}{c|}{\begin{tabular}[c]{@{}c@{}}Power \\ (kWh)\end{tabular}} & \multicolumn{1}{c}{Time} & \multicolumn{1}{c|}{\begin{tabular}[c]{@{}c@{}}Power \\ (kWh)\end{tabular}}& 

        \multicolumn{1}{c}{Time}& \multicolumn{1}{c|}{\begin{tabular}[c]{@{}c@{}}Power\\ (kWh)\end{tabular}} & \multicolumn{1}{c}{Time}& \multicolumn{1}{c|}{\begin{tabular}[c]{@{}c@{}}Power\\ (kWh)\end{tabular}} & \multicolumn{1}{c}{Time}& \multicolumn{1}{c|}{\begin{tabular}[c]{@{}c@{}}Power\\ (kWh)\end{tabular}} &
        
        \multicolumn{1}{c}{RPi 4} & \multicolumn{1}{c}{Nano} & Orin & \multicolumn{1}{c}{RPi 4} & \multicolumn{1}{c}{Nano} & Orin\\ \midrule
        
        FEMNIST & CNN & 0.1 MB&         20s& 0.0006& 0.8s& 0.0001& 56s& 0.0451& 31s& 0.0352& 11s& 0.0059& 791 & 438& 579& 35000 & 19375& 25625 \\
        BLOND & CNN & 0.1 MB&           20s& 0.0003& 0.8s& 0.0001& 65s& 0.0530& 23s& 0.0269& 11s& 0.0059 & 918 & 325& 155& 40625 & 14375& 6875\\
        BLOND & LSTM& 0.2 MB&           20s& 0.0007& 0.8s& 0.0001& 80s&0.0652& 21s& 0.0243& 13s& 0.0074& 565 & 148& 92 & 25000 & 6563 & 4063\\
        BLOND & ResNet& 0.4 MB&         30s& 0.0016& 1.2s& 0.0002& 588s& 0.5197& 36s& 0.0417& 15s& 0.0086& 2076& 127& 53 & 91875 & 5625 & 2344\\
        BLOND & DenseNet& 1.0 MB&       70s& 0.0041& 2.8s& 0.0006& 586s& 0.5125& 36s& 0.0416& 16s& 0.0086& 828& 51& 23& 36625& 2250 & 1000\\
        Shakespeare & LSTM& 3.2 MB&     240s& 0.0134& 9.6s& 0.0021& 981s& 0.9172& 59s& 0.0724& 19s&  0.0103& 433 & 26 & 8& 19160 & 1152 & 371 \\
        SAMSum& FLAN-T5 Small & 308 MB& 21800s& 1.3124& 872s& 0.2008& OOM& OOM& OOM& OOM& 324s& 2.1492& OOM & OOM& 1.5& OOM & OOM& 65.7\\ \bottomrule \bottomrule
        \end{tabular}
    }
\end{table*}
\vspace{-8pt}
\subsection{Client behavior}
\label{sec:system-robustness}

\textbf{Existing FL strategies work well with unreliable clients}. Unreliability is a core challenge for edge computing systems. 
Especially for FL workloads, this bears the potential for significant information loss whenever model updates are not \newline\vspace{30\baselineskip}\newline transmitted to the server, regardless of the root cause. 
As embedded devices do not have power redundancy, run in sub-optimal environments w.r.t. heat dispersion, and may suffer from outside damages, they are considered unreliable by nature. 
Our experiments show that existing state-of-the-art FL strategies work well in systems with unreliable clients, i.e., high dropout rates.
However, for each dataset, we see that one FL strategy always consistently performs best. This suggests that the FL strategy choice overall depends on the dataset, not the model.
Yet, we also identify FedAdaGrad and FedAdam as particularly sensitive to client dropout. 
Thus, careful strategy selection and federated hyperparameter optimization are critical for systems that involve unreliable clients. 

\subsection{Differential Privacy}
\label{sec:dp-experiments}
\textbf{Heterogeneous client behavior has significant negative effects on the model quality of differentially private workloads}. Since DP is an integral component of FL workloads, the question arises: how does it change the model quality with varying client reliability levels? 
Interestingly, the model quality decreases significantly when introducing client dropout in a system, and the negative effects across strategies become significant at a low DP level. 
Since we use a server-side user-level DP algorithm, we adjust the noise level based on the model updates received per training round. 
This provides appropriate privacy guarantees. 
Yet, the level of noise we have to add to cover for failing clients reduces the model quality significantly  (\Cref{tab:compound-effects}).
For instance, if we want to provide a loose privacy budget of $\epsilon = 8$ on the BLOND dataset and suffer from a high client dropout rate ($p = 0.5$), the accuracy almost halves compared to a system without client dropouts ($p = 0$).
As such, allowing re-training with another client in a training round or accounting for late arrivals could mitigate the effects DP has on the model quality. 
Also, the evaluation of DP in unreliable systems is a strong indicator of whether deploying an FL workload into an edge computing system w.r.t. to the estimated model quality in relation to the privacy requirements is worthwhile. 
However, training quality and efficiency are equally relevant for an effective deployment.

\subsection{Communication Efficiency}
\label{sec:comms-efficiency}

\textbf{Granularity helps quantify the practical utility in relation to the network when deploying FL workloads in edge computing systems}. Before discussing the hardware characteristics of diverse embedded platforms, the question is whether it is even worthwhile deploying an FL workload to them under given network conditions.

 With larger models, FL workloads become more communication intensive as we have to send more model weights. From a practical perspective, quantifying whether it's worthwhile to consider including embedded devices in a system that runs an FL workload is essential. For all of our state-of-the-art FL workloads, $G$ is significantly above 1 (\Cref{tab:granularity-network}).
This indicates the high suitability of such small models to be trained as an FL workload, regardless of the communication technology. 
This is particularly useful for highly specialized tasks like object detection. 
Yet, for more generalizing tasks like in the NLP space, we require larger models to run on embedded devices and be trained at the edge. For larger models like FLAN-T5 Small, we see $G = 1.5$. With $G$ close to 1, the practical utility is limited as communication takes approximately as much time as computation. Beyond the computation/communication trade-off, $G$ is also suitable for quantifying the net effect of communication optimization methods for an FL workload.
Further, reliability is a major concern in edge computing systems with embedded clients. We want to spend as little time communicating as possible to get the model updates to the server and not risk failures during long communication times. However, computational performance and efficiency are key challenges as well.

\subsection{Energy Efficiency}
\label{sec:energy-eficiency-analysis}
\textbf{Energy efficiency is a well-suited indicator for computational bottlenecks on embedded devices and does not require extensive client monitoring that could potentially infringe privacy}. As we will see in \Cref{sec:hardware-heterogeneity}, the measurements of local step times on clients unveil inefficiencies and scalability limits. Yet, micro-benchmarks are often difficult to facilitate when having a variety of devices in a system. While micro-benchmarks are useful for an in-depth exploration of bottleneck root causes on a client, they require significant efforts and interfere with the DL training process. So, how can we identify bottlenecks without interfering with the client's training process?

\begin{figure*}
    \centering
    \begin{subfigure}[!ht]{0.68\textwidth}
        \raisebox{-\height}{\includegraphics[height=4.8cm]{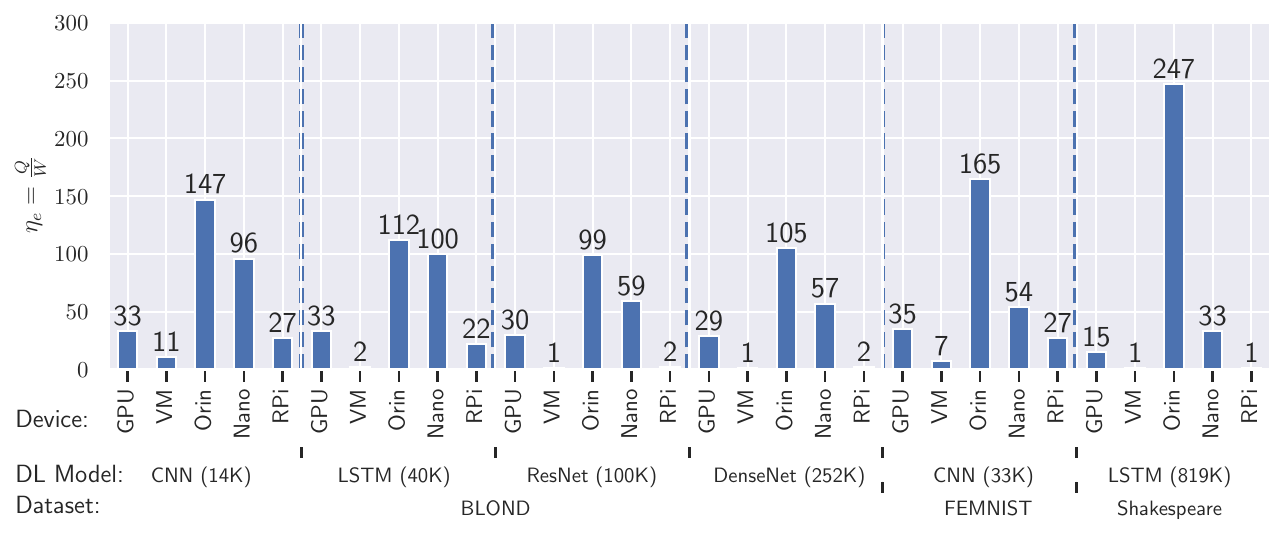}}
        \caption{Device comparison}
        \label{fig:energy-devices}
    \end{subfigure}
    \begin{subfigure}[!ht]{0.23\textwidth}
        \raisebox{-\height}{\includegraphics[height=4.1cm]{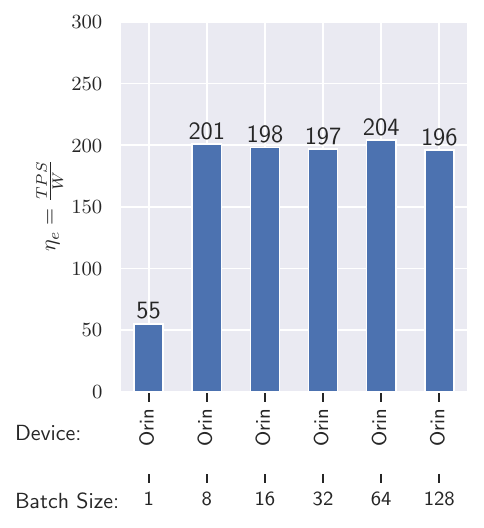}}
        \caption{Scaling batch size with the FLAN-T5 Small (80M) transformer}
        \label{fig:energy-t5-small}
    \end{subfigure}

    \caption{Energy efficiency measured across datasets and DL models (\# parameters).\protect\footnotemark[5] Results are measured over one epoch of training. The Orins train with a minibatch size of 256 across all models. Higher values are better.}
    \label{fig:sample-efficiency}
\end{figure*}

We find energy efficiency to be a well-suited estimator for the suitability of a device for a given task and also for the identification of bottlenecks (\Cref{fig:sample-efficiency}). The main benefit of energy efficiency is that it measures the throughput per Watt of power consumption, two available metrics without interference with the training process, i.e., they are convenient to measure. 
The device comparison shows the significant advancements in hardware for FL workloads on the edge (\Cref{fig:energy-devices}). It is also an indicator that we should look more at models with multi-million parameter models for FL workloads. The Orins have proven to become more energy efficient with increasing parameter size and outperform all other embedded device platforms as well as data center resources for state-of-the-art FL workloads. 
We identify the computational capabilities of the Orins by running the 80M parameter FLAN-T5 model for the SAMSum text summarization task (\Cref{fig:energy-t5-small}). By scaling the batch size, we quickly discover that the energy efficiency stagnates at a batch size of 8 and does not increase any further. 
The reason is the significant memory bottleneck we uncover with the micro-benchmark (see \Cref{sec:hardware-heterogeneity}). Yet, this bottleneck is immediately visible without interfering with the training process itself. 
When monitoring embedded devices in practice, it is a necessity to have easy-to-interpret metrics for quick responses to inefficiencies.
As such, energy efficiency covers two integral aspects of deploying FL workloads in edge computing systems. First, we can quantify the suitability of an embedded device for the deployment of an FL workload. 
Second, computational limitations become instantaneously visible for a remote operator of an FL system. This is particularly important since an operator cannot access the devices. 

\subsection{Hardware Heterogeneity}
\label{sec:hardware-heterogeneity}

\begin{figure*}[!ht]
    \centering
    \begin{subfigure}[!ht]{0.68\textwidth}
        \raisebox{-\height}{\includegraphics[height=4.8cm]{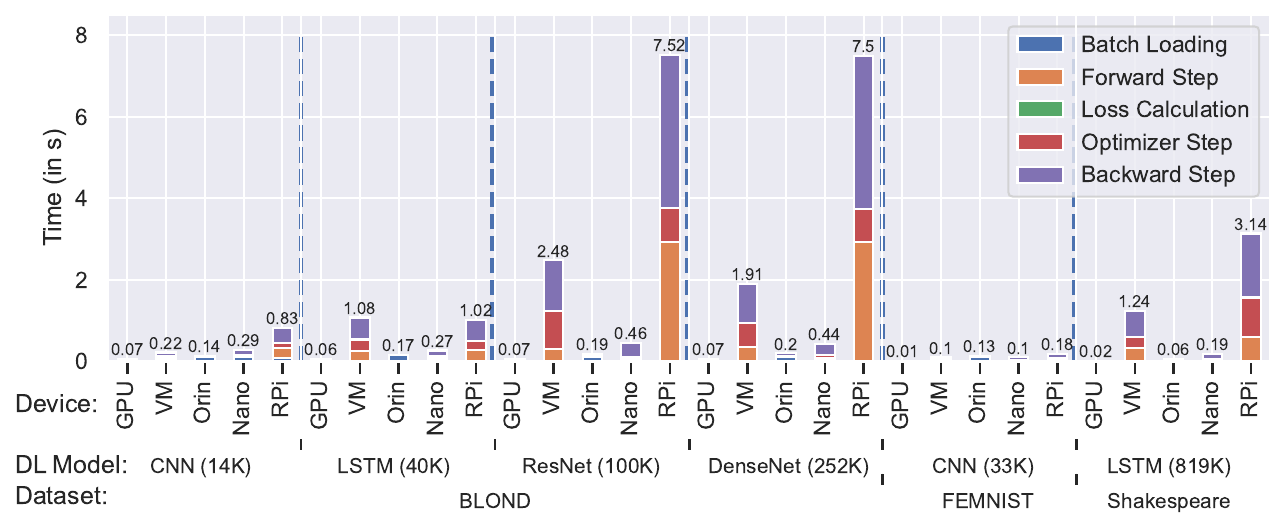}}
        \caption{Device comparison}
        \label{fig:microbench-devices}
    \end{subfigure}
    \begin{subfigure}[!ht]{0.26\textwidth}
        \vspace{4pt}
        \raisebox{-\height}{\includegraphics[height=4.5cm]{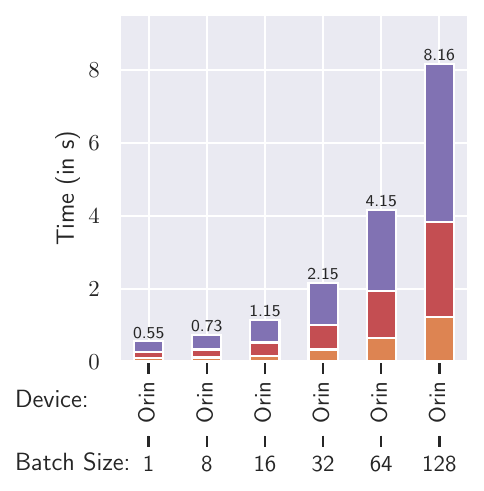}}
        \caption{Scaling batch size with the FLAN-T5 Small (80M) transformer}
        \label{fig:microbench-t5-small}
    \end{subfigure}

    \caption{Training times for different device types and model sizes over one minibatch. We further scale the minibatch size for the FLAN-T5 transformer model to showcase edge-specific bottlenecks. Next to the model name, we report the model parameters. Lower is better.}
    \label{fig:microbench}
\end{figure*}

\textbf{Contrary to current trends in high-performance DL, scaling the batch size on embedded devices does not lead to greater computational efficiencies}. For our benchmark, embedded hardware forms the very center of attention since it ultimately depends on the devices deployed in the field how well an FL workload will run (\Cref{fig:microbench-devices}). Therefore, our micro-benchmark provides a comprehensive baseline for the suitability of device/workload combinations. In systems with legacy hardware, as it is often found in industrial systems with long product lifecycles, the benchmark reveals their limited utility for FL workloads. 
This is especially evident for workloads that entail models with several million parameters since scaling the minibatch size beyond 8 samples does not yield performance benefits. 
As we double the minibatch size, the runtime also doubles.


        

This originates from the limited memory bandwidth of embedded devices and first affects the optimizer step function, which requires a significant amount of interactions between processor and memory.
Interestingly, for the models larger than 100K parameters, the forward step also grows exponentially. This operation does not require many interactions between computing and memory, only batch loading once per batch to send the data through the DL model. One could think this might be an I/O bottleneck originating from the disk. Yet, we run the same SD cards on the Nanos and RPis. The Nanos do not show this behavior in any use case.

The Orin platform, as the most recent embedded device type, yields strong results throughout our experiments. Yet, they are made for significantly larger workloads involving DL models with several million parameters. Therefore, we study their behavior when fine-tuning the FLAN-T5 Small model with different batch sizes (\Cref{fig:microbench-t5-small}). We see linearly growing step times, which is a strong sign of a bottleneck, as one would normally expect to see a logarithmically increasing step time with scaling batch size. The bottleneck originates from the limited memory bandwidth of the Orins (204 GB/s) vs. modern data center DL accelerators (e.g., the A6000 with 768 GB/s). Therefore, federated hyperparameter optimization is not only a matter of model quality but must also consider the hardware limitations.

\revision{
Our profiling experiments with the FLAN-T5 Small model on the Orins and data center GPUs show two main bottlenecks. 
The profiling was done for an identical workload used across the different hardware with a minibatch size of 128 samples.

We start with CPU-related bottlenecks. 
The performance difference between the ARM-based Orins and the x86-based CPU in our GPU server becomes evident in the \texttt{opt.step()}, executed on each device's CPU. 
Overall, the \texttt{opt.step()} takes $4\times$ longer on the Orins, which is primarily rooted in an \texttt{aten::\_foreach\_add} operation that performs an element-wise addition of tensors.
Within this operation, there are noteworthy differences in how much time is spent on the actual computation and how much is used for data movement from and to the main memory. 
The Orins spend $83\%$ of the operation time on data movement, while the GPU server only needs $77\%$. 
This originates from a different handling of the \texttt{aten::copy\_} operation (to move data from GPU to CPU memory) on the two devices. 
Since the Orins have shared memory for the CPU and GPU, data is assigned to either device via a context switch, while on the GPU server, data is moved between the main memory and GPU memory. 
One would expect a context switch to be faster than copying data from one memory to another. 
However, we observe that the context switch takes between $5 - 8 \mu s $, regardless of the data size, while the data movement on the GPU server depends on the amount of data transferred, taking a maximum of $1 \mu s$.
Overall, this requires the Orins to spend more time on data movement.

The second major bottleneck is located on the GPU during the forward step. 
At a high level, we find that the GPU on the Orins is permanently busy with operations, while the GPU server has notable idle times.
We identify matrix multiplications on the GPU as the main driver for the longer processing times on the Orins. 
This is a sign of a computational bottleneck. 
Specifically, \texttt{ampere\_sgemm\_128x128\_nn}, a compute-intensive operation, stands out since it takes up to $32\times$ longer on the Orins than on the A6000 (11ms vs. $323\mu s$).
This is rooted in the significantly higher number of tasks (warps) per GPU streaming multiprocessor (SM) on the Orins. 
We see 1200 warps per SM on the Orins, while there are only 37 on the A6000.
As such, the Orins have to process $32\times$ more tasks per SM than the A6000.
At the same time, we also observe longer memory operation times. 
Immediately preceding the matrix multiplication, the \texttt{direct\_copy\_kernel\_cuda} data movement operator from CPU to GPU memory takes up to $5.9\times$ longer on the Orins vs. the A6000. 

Overall, we observe a major limitation of the Orins when it comes to computational intensity and data movement between CPU and GPU, even though the device has a shared memory architecture.
}

\noindent As such, the micro-benchmark design \revision{paired with detailed profiling results} aids not only in identifying the right device, workload, and hyperparameter combinations but also reveals hardware bottlenecks on the kernel level. Further, it helps detect defective parts and can be used as a predictive monitoring element to improve system stability, even though the clients may not be directly accessible.

\footnotetext[5]{We use estimates from SelfWatts \cite{Fieni2021} for approximating the power consumption of VMs.}

    \section{Related Work}
    \label{sec:related-work}
    
\textbf{Edge computing}. For Edge Computing systems, there are numerous benchmarking tools. Edge devices have long been regarded as a data pass-through gateway rather than data processing devices and, therefore, have been evaluated in the context of data transfer capacities and reliability \cite{Kruger2014}. 
In 2019, \citet{McChesney2019} introduced DeFog, a benchmark including DL inference workloads on embedded devices with YOLO models. 
Yet, these benchmarks do not include system dependencies between clients and a server as we would find in FL systems, where training progress relies on client participation.
\citet{Varghese2022} provide a comprehensive overview of a wide range of edge computing benchmarks, but none is targeted to FL.

\revision{
\begin{table}[!ht]
    \centering
    \caption{\revision{Comparison of FLEdge with other FL benchmarks.}}
    \label{tab:benchmark_comparison}
    \revision{
        \resizebox{0.48\textwidth}{!}{
            \begin{threeparttable}
                \begin{tabular}{@{}l|ll|ccc|cccc@{}}
    \toprule
              &     & \multicolumn{1}{c}{Primary}                                            & \multicolumn{3}{|c|}{ML Domains}                                                               & \multicolumn{4}{c}{Analysis Dimensions}                                                                                                                                                   \\
    Benchmark & Year & Eval. Purpose                  & \multicolumn{1}{c}{CV} & \multicolumn{1}{c}{NLP} & \multicolumn{1}{c|}{NILM} & \multicolumn{1}{c}{(1)} & \multicolumn{1}{c}{(2)} & \multicolumn{1}{c}{(3)} & \multicolumn{1}{c}{(4)} \\ \midrule
    LEAF               & 2018          & Aggregation                             & \checkmark                               & \checkmark                                &                                    &                                            &                                                        &                                              &                                                  \\
    FedML              & 2020          & Aggregation                             & \checkmark                               & \checkmark                                &                                    &                                            &                                                        &                                              &                                                  \\
    Flower             & 2020          & Aggregation                             & \checkmark                               & \checkmark                                &                                    & \checkmark                                          &                                                        &                                              &                                                  \\
    FederatedScope     & 2022          & Personalization            & \checkmark                               & \checkmark                                &                                    & \checkmark                                          &                                                        &                                              &                                                  \\
    FedScale           & 2022          & Scalability                            & \checkmark                               & \checkmark                                &                                    & \checkmark                                           &                                                        &                                              &                                                  \\
    \textbf{FLEdge}    & 2024          & FL Clients & \checkmark                               & \checkmark                                & \checkmark                                  & \checkmark                                          & \checkmark                                                      & \checkmark                                            & \checkmark                                                \\ \bottomrule \bottomrule
\end{tabular}
                \begin{tablenotes}
                        \small
                        \item Analysis Dims.: (1) Data Security, (2) Dedicated Edge Deployment, (3) Client Behavior, (4) Client Capabilities    
                \end{tablenotes}
            \end{threeparttable}
        }
    }
\end{table}
}

\revision{
    \textbf{FL benchmarks}. 
    LEAF \cite{caldas1} and FedML \cite{He2020} introduce two benchmarks that focus on the evaluation of FL optimization algorithms with data heterogeneity. 
    Flower \cite{beutel2020} and FederatedScope \cite{Xie2022_FederatedScope} extend FL benchmark capabilities with private computing by using $(\epsilon, \delta)$-DP as well as cryptographic aggregation. 
    FederatedScope also enables benchmarking of personalized FL. 
    FedScale \cite{Lai2021} evaluates the scalability characteristics of FL systems.
    With FLEdge, we complement the benchmarking landscape for FL applications with client capability and behavior analysis tools that integrate with previous benchmarking works. Further, we are the first to design a benchmark that is entirely based on dedicated hardware. Our study emulates wide-area networking and client behavior to enable systematic analyses (\Cref{tab:benchmark_comparison}). 
}

\textbf{Real-world FL workloads}. It is particularly challenging to evaluate FL workloads in real-world edge computing systems as devices are often not accessible for research. 
FedScale \cite{Lai2021} shed light on the high-level performance characteristics of two mobile device platforms for FL workloads. 
The same is true for FS-Real \cite{Chen2023_FSReal}, which is based on FederatedScope. 
It presents a runtime optimized for mobile devices running Android.
FLINT by \citet{wang2023flint} presents an approach to integrating FL workloads into existing ML landscapes and discusses the effects of scaling DL models beyond cloud resources onto mobile devices on the network edge. 
While Wang et al. discuss the utility of hardware analytics for task scheduling in FL systems, they outline the open necessity for in-depth hardware and energy analysis due to the wide variety of devices being available for FL workload deployment.
As such, there is still a lack of real-time analysis of FL workloads on low-power devices and missing guidance on in-depth performance differences between platforms, especially since the landscape of DL-accelerated embedded devices is growing rapidly.

With \papername, we expand on hardware-centric research by conducting in-depth evaluations of compute performance across various embedded device generations and the role of energy in FL workloads.
To address the distinctive characteristics of client behavior in edge computing systems, we extensively study the effects on DP FL workloads.

    \vspace{-12pt}
    \section{Lessons Learned}
    \label{sec:lessons-learned}
    We summarize what we learned from studying FL from a hardware centric perspective. Findings relate to client behavior that we must account for in the future, improving FL applications based on the hardware specifications we find on the network edge and what workloads are beneficial to be deployed to embedded hardware.

\textbf{Client participation patterns in FL are determinant for reliable training that entails $(\epsilon, \delta)$-DP}.
We find that client participation significantly negatively affects the training performance of FL workloads \revision{(\Cref{sec:system-robustness})}. 
While we provide an approach to adjusting DP noise levels based on the number of clients that submit model updates, client dropouts lead to higher noise levels to provide the same privacy guarantees.
This degrades the training performance and requires more time and resources to achieve the same performance as in FL systems that do not employ DP.
\revision{Yet, to better understand how $(\epsilon, \delta)$-DP can be applied in practice, it is imperative to further study what $\epsilon$ levels can be considered acceptable.}
At the same time, it is a priority to understand realistic client participation statistics and build middleware that can account for unreliable clients \cite{wang2024a}. 

\textbf{Larger ML models benefit the scalability}.
With larger models, we spend more time on computation and less on communication \revision{(\Cref{sec:comms-efficiency})}.
Thus, the overall FL process becomes more efficient since more time is spent learning from client data.
Thus, workload and hardware fit can help drive overall system efficiency, carefully balancing training speed and power draw.

\revision{
\textbf{Context switches for data assignment between CPU and GPU is a major limiting factor on embedded hardware}. 
Despite the unified memory architecture on state-of-the-art embedded hardware, context switches between CPU and GPU memory are time-consuming compared to the physical data movement operation on data center GPUs (\Cref{sec:hardware-heterogeneity}). 
This can be a major cost driver for FL applications in edge computing systems and should be considered when deciding on data movement frequency.
}

    \section{Conclusions}
    \label{sec:discussion-conclusion}
    In this paper, we present FLEdge, a hardware-centric benchmark for deploying FL workloads on the edge. We study hardware diversity, efficiency, and robustness. 
Our experimental results show the significance of hardware diversity for the training performance of FL workloads, along with the important role of network conditions. We show the limits of current state-of-the-art embedded devices and the effects of DL hyperparameters on training efficiency. The minibatch size, which is a key factor in scaling DL workloads in data centers, shows very limited effects for FL workloads on the edge.
Our experiments on energy efficiency introduce an early detector for computational bottlenecks that does not require interference with the training process. As such, it is suitable for further evaluating the suitability of energy efficiency as an optimization lever for FL workloads. 
While our experiments on heterogeneous client behavior show the robustness of state-of-the-art FL strategies, we identify a particular sensitivity of DP FL workloads to client failures, which is a critical factor when deploying FL workloads that involve sensitive data. 
Overall, FLEdge, our hardware-centric benchmark, contributes to the practical considerations required to deploy FL workloads to edge computing systems, and we hope to further spur research toward FL middleware that can control both on-client and global learning efficiency.

    \newpage
    
    \section*{Acknowledgements}
    This work is partially funded by the German Federal Ministry of Economic Affairs and Climate Action (Grant: 16KN085729) and the Bavarian Ministry of Economic Affairs, Regional Development and Energy (Grant: DIK0446/01).
    We would like to thank the PANDORA project (\url{https://pandora-heu.eu/}) for our fruitful discussions. 
    
    {
        \balance
        \bibliographystyle{ACM-Reference-Format}
        \bibliography{main}
    }
    
\end{document}